\documentclass[10pt,twocolumn,letterpaper]{article}

\usepackage{iccv}
\usepackage{times}
\usepackage{epsfig}
\usepackage{graphicx}
\usepackage{amsmath}
\usepackage{amssymb}

\usepackage[ruled,vlined]{algorithm2e}
\usepackage{booktabs}       
\usepackage{amsfonts}       
\usepackage{nicefrac}       
\usepackage{microtype}  

\usepackage{multirow}
\usepackage{array}
\usepackage{rotating} 
\usepackage{wrapfig}
\usepackage{tabularx}
\usepackage{pifont}
\usepackage{bbm}
\usepackage{color, colortbl}
\definecolor{LightCyan}{rgb}{1, 0.898, 0.8}

\usepackage[pagebackref=true,breaklinks=true,letterpaper=true,colorlinks,bookmarks=false]{hyperref}

\iccvfinalcopy 

\def\iccvPaperID{7185} 

\ificcvfinal\fi

\begin{document}
	
	\title{Exploring Robustness of Unsupervised Domain Adaptation in Semantic Segmentation}
	
	\author{Jinyu Yang\textsuperscript{1}, Chunyuan Li\textsuperscript{1}, Weizhi An\textsuperscript{1}, Hehuan Ma\textsuperscript{1}, Yuzhi Guo\textsuperscript{1}, Yu Rong\textsuperscript{2}, Peilin Zhao\textsuperscript{2}, Junzhou Huang\textsuperscript{1}\\
	\textsuperscript{1}University of Texas at Arlington\\
	\textsuperscript{2}Tencent AI Lab\\
	}
	
	\maketitle
	\ificcvfinal\thispagestyle{empty}\fi
	
	\begin{abstract}
		Recent studies imply that deep neural networks are vulnerable to adversarial examples---inputs with a slight but intentional perturbation are incorrectly classified by the network. Such vulnerability makes it risky for some security-related applications (e.g., semantic segmentation in autonomous cars) and triggers tremendous concerns on the model reliability. For the first time, we comprehensively evaluate the robustness of existing UDA methods and propose a robust UDA approach. It is rooted in two observations: (i) the robustness of UDA methods in semantic segmentation remains unexplored, which pose a security concern in this field; and (ii) although commonly used self-supervision (e.g., rotation and jigsaw) benefits image tasks such as classification and recognition, they fail to provide the critical supervision signals that could learn discriminative representation for segmentation tasks. These observations motivate us to propose adversarial self-supervision UDA (or ASSUDA) that maximizes the agreement between clean images and their adversarial examples by a contrastive loss in the output space. Extensive empirical studies on commonly used benchmarks demonstrate that ASSUDA is resistant to adversarial attacks.
	\end{abstract}
	
	\section{Introduction}
	Semantic segmentation aims to predict semantic labels of each pixel in the given images, which plays an important role in autonomous driving \cite{luc2017predicting} and medical diagnosis \cite{raju2020co}.
	However, pixel-wise labeling is extremely time-consuming and labor-intensive.
	For instance, 90 minutes are required to annotate a single image for the Cityscapes dataset \cite{cordts2016cityscapes}.
	Although synthetic datasets \cite{richter2016playing,ros2016synthia} with freely-available labels provide an opportunity for model training, the model trained on synthetic data suffers from dramatic performance degradation when applying it directly to the real data of interest.
	
	Motivated by the success of unsupervised domain adaptation (UDA) in image classification, various UDA methods for semantic segmentation are recently proposed.
	The key idea of these methods is to learn domain-invariant representations by minimizing marginal distribution distance between the source and target domains \cite{hoffman2016fcns}, adapting structured output space \cite{tsai2018learning,chen2018road}, or reducing appearance discrepancy through image-to-image translation \cite{hoffman2017cycada,zhang2018fully,li2019bidirectional}.
	Another alternative is to explicitly explore the supervision signals from the target domain through self-training.
	The key idea is to alternatively generate pseudo labels on target data and re-train the
	model with these labels.
	Most of the existing state-of-the-art UDA methods in semantic segmentation rely on this strategy and demonstrate significant performance improvement.  \cite{zou2018unsupervised,li2019bidirectional,huang2020contextual,pan2020unsupervised,wang2020differential,kim2020learning,yang2020fda,yang2020label,pan2020two}.
	However, the limitation of self-training-based UDA methods lies in that pseudo labels are noisy and less accurate, giving rise to severe label corruption problems \cite{patrini2017making,zhang2018generalized}.
	Recent studies further prove that with high degrees of label corruption, models tend to overfit the misinformation in the corrupted labels \cite{zhang2018generalized,hendrycks2019using}.
	As a consequence, performance improvement cannot be guaranteed by simply increasing the re-training rounds.
	
	Another critical issue of UDA methods in semantic segmentation is that they are possibly vulnerable to adversarial attacks. 
	In other words, the performance of a UDA model may dramatically degrade under an unnoticeable perturbation.  
	Unfortunately, the robustness of UDA methods remains largely unexplored.
	With the increasing applications of UDA methods in security-related areas, the lack of robustness of these methods leads to massive safety concerns.
	For instance, even small-magnitude perturbations on traffic signs can potentially cause disastrous consequences to autonomous cars \cite{eykholt2018robust,sitawarin2018darts}, such as life-threatening accidents.
	
	Self-supervised learning (SSL) aims to learn more transferable and generalized features for vision tasks (\eg, classification and recognition) \cite{dosovitskiy2015discriminative,gidaris2018unsupervised,he2020momentum,chen2020simple}. 
	Key to SSL is the design of pretext tasks, such as rotation prediction, selfie, and jigsaw, to obtain self-derived supervisory signals on unlabeled data.
	Recent studies reveal that SSL is effective in improving model robustness and uncertainty \cite{hendrycks2019using}.
	However, the commonly used pretext tasks fail to provide critical supervision signals for segmentation tasks where fine-grained features are required \cite{zhan2017mix}.
	
	In this paper, we first perform a comprehensive study to evaluate the robustness of existing UDA methods in semantic segmentation.
	Our results first reveal that these methods can be easily fooled by small perturbations or adversarial attacks.
	We, therefore introduce a new UDA method known as ASSUDA to robustly adapt domain knowledge in urban-scene semantic segmentation.
	The key insight of our method is to leverage the regularization power of adversarial examples in self-supervision.
	Specifically, we propose the adversarial self-supervision that maximizes the agreement between clean images and their adversarial examples by a contrastive loss in the output space.
	The key to our method is that we use adversarial examples to (i) provide fine-grained supervision signals for unlabeled target data, so that more transferable and generalized features can be learned and (ii) improve the robustness of our model against adversarial attacks by taking advantages of both adversarial training and self-supervision.
	
	Our main contributions can be summarized as: (i) To the best of our knowledge, this paper presents the first systematic study on how existing UDA methods in semantic segmentation are vulnerable to adversarial attacks. 
	We believe this investigation provides new insight into this area.
	(ii) We propose a new UDA method that takes advantage of adversarial training and self-supervision to improve the model robustness.
	(iii) Comprehensive empirical studies demonstrate the robustness of our method against adversarial attacks on two benchmark settings, \ie, "GTA5 to Cityscapes" and "SYNTHIA to Cityscapes".
	
	\section{Related Work}
	\paragraph{Unsupervised Domain Adaptation}
	Unsupervised domain adaptation (UDA) refers to the scenario where no labels are available for the target domain.
	In the past few years, various UDA methods are proposed for semantic segmentation, which can be mainly summarized as three streams: 
	(i) adapt domain-invariant features by directly minimizing the representation distance between two domains \cite{hoffman2016fcns,zhu2018penalizing}, 
	(ii) align pixel space through translating images from the source domain to the target domain \cite{hoffman2017cycada,murez2017image}, and
	(iii) align structured output space, which is inspired by the fact that source output and target output share substantial similarities in terms of structure layout \cite{tsai2018learning}.
	However, simply aligning cross-domain distribution has limited capability in transferring pixel-level domain knowledge for semantic segmentation.
	To address this problem, the most recent studies integrate self-training into existing UDA frameworks and demonstrate the state-of-the-art performance \cite{zou2018unsupervised,li2019bidirectional,huang2020contextual,pan2020unsupervised,wang2020differential,kim2020learning,yang2020fda,yang2020label,pan2020two}. 
	However, the most obvious drawback of such a strategy is that the generated pseudo labels are noisy and the model tends to overfit the misinformation in the corrupted labels. 
	
	Our method instead resorts to self-supervision by integrating contrastive learning into existing UDA methods.  
	This strategy demonstrates two advantages, \ie, (i) provides supervision for the target domain, which is proved to be robust to the label corruption, and (ii) encourages the model to learn more transferable and robust features.
	Another major difference is that our method mainly focuses on improving its robustness against adversarial attacks, which is overlooked by existing UDA methods.
	
	\begin{figure*}[t]
		\begin{center}
			\includegraphics[width=1.0\linewidth]{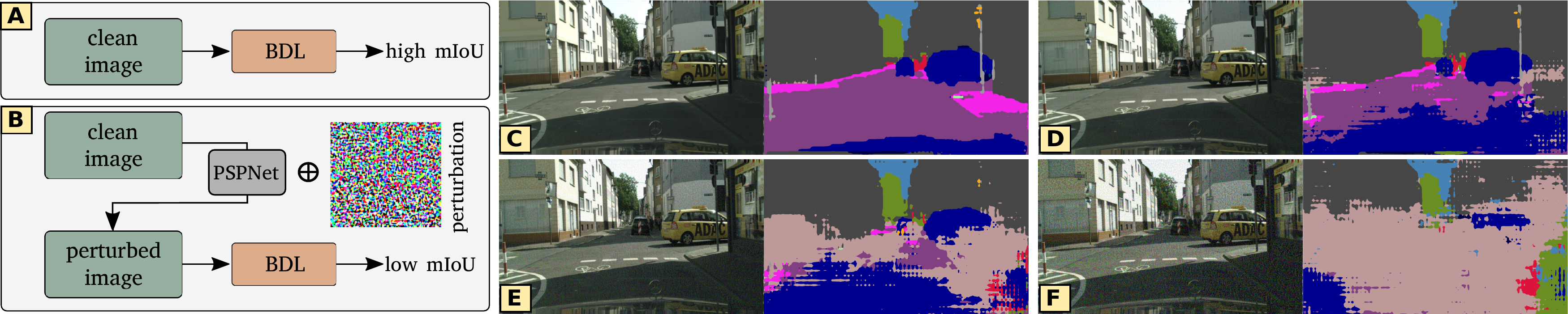}
		\end{center}
		\caption{Robustness study of BDL \cite{li2019bidirectional} on "GTA5 to Cityscapes" with backbone ResNet101. (A) the traditional paradigm uses clean test data to evaluate the performance of BDL; (B) we use PSPNet as the surrogate model to generate perturbed test data which are then used to evaluate BDL; (C) a clean image and its segmentation output predicted by BDL; (D), (E), and (F) indicate the perturbed images of (C) with $\epsilon=0.1$, $\epsilon=0.25$, and $\epsilon=0.5$, respectively, along with their BDL predictions. Although the perturbations are unnoticeable, they can easily deceive BDL, resulting in dramatic performance degradation.}
		\label{fig:attack_BDL_fig}
		\vspace{-0.2in}
	\end{figure*}

	\vspace{-10pt}
	\paragraph{Self-supervised Learning}
	Self-supervision aims to make use of massive amounts of unlabeled data through getting free supervision from the data itself.
	This is typically achieved by training self-supervised tasks (a.k.a., pretext tasks) through two paradigms, \ie, pre-training \& fine-tuning or multi-task learning.
	Specifically, the pre-training \& fine-tuning first performs pre-training on the pretext task, then fine-tunes on the downstream task.
	In contrast, multi-task learning optimizes the pretext task and the downstream task simultaneously.
	Our method falls into the latter, where the downstream task is to predict the segmentation labels of the target domain.
	To learn transferable and generalized features through self-supervision, it is essential to design pretext tasks that are tailored to the downstream task. 
	Commonly used pretext tasks include exemplar \cite{dosovitskiy2015discriminative}, rotation \cite{gidaris2018unsupervised}, predicting the relative position between two random patches \cite{doersch2015unsupervised}, and jigsaw \cite{noroozi2016unsupervised}.
	Motivated by this, recent UDA methods introduce self-supervision into segmentation adaptation to learn domain invariant feature representations \cite{xu2019self,sun2019unsupervised}.
	Although commonly used pretext tasks contribute to cross-domain feature alignment, these tasks have limited ability in learning fine-grained representations that are essential in semantic segmentation.
	
	By contrast, this paper proposes to use adversarial examples to build pretext tasks.
	Different from \cite{chen2020simple} that perform contrastive learning in the latent space, we maximize agreement between each sample and its adversarial example via a contrastive loss in the output space.
	Therefore, (i) our method is encouraged to learn more transferable features which are domain invariant and fine-grained, and (ii) the trained model is more robust to label corruption and adversarial attacks.
	
	\vspace{-10pt}
	\paragraph{Adversarial Attacks}
	Previous studies reveal that adversarial attacks are commonly observed in machine learning methods such as SVMs \cite{biggio2012poisoning} and logistic regression \cite{mei2015using}.
	Recent publications suggest that neural networks are also highly vulnerable to adversarial perturbations \cite{szegedy2013intriguing,goodfellow2014explaining}. 
	Even worse, adversarial attacks are proven to be transferable across different models \cite{tramer2017space}, i.e., the adversarial examples generated to attack a specific model are also harmful to other models.
	To fully understand adversarial attacks in deep neural networks (DNNs), considerable attention is received in the past few years. 
	Specifically, \cite{goodfellow2014explaining} proposes a fast gradient sign method (FGSM) to efficiently generate adversarial examples with only one gradient step. 
	DeepFool \cite{moosavi2016deepfool} generates minimal perturbations by iteratively linearizing the image classifier.
	By utilizing the differential evolution, \cite{su2019one} enables us to generate one-pixel adversarial perturbations to accurately attack DNNs.
	
	Unlike the aforementioned studies that focus on effectively creating adversarial attacks, our method uses adversarial examples to build pretext tasks for UDA models, and in turn to improve the model robustness.
	This is motivated by the fact that a clean image and its adversarial example should have the same segmentation output.
	Therefore, we can get supervision for free and encourage our method to learn discriminative representation for segmentation tasks.
	
	\begin{table}
		\caption{Performance comparison of BDL that is evaluated on clean test data VS. perturbed test data. Three sets of perturbed data are generated with $\epsilon=0.1$, $\epsilon=0.25$, and $\epsilon=0.5$, respectively.}
		\label{table:attack_BDL}
		
		\footnotesize
		\setlength\tabcolsep{9pt}
		\begin{center}
			\begin{tabularx}{.45\textwidth}{ c|c|c|c @{} }
				\toprule
				Base & $\epsilon$ & GTA5 to City & SYNTHIA to City \\
				\midrule
				\multirow{3}{*}{VGG16}
				& 0.1 & 41.3 $\rightarrow$ 30.5 & 39.0 $\rightarrow$ 29.3 \\
				& 0.25 & 41.3 $\rightarrow$ 14.6 & 39.0 $\rightarrow$ 13.6 \\
				& 0.5 & 41.3 $\rightarrow$ 7.10 & 39.0 $\rightarrow$ 5.90 \\
				
				\midrule
				\multirow{3}{*}{ResNet101}
				& 0.1 & 48.5 $\rightarrow$ 36.2 & 51.4 $\rightarrow$ 41.2 \\
				& 0.25 & 48.5 $\rightarrow$ 19.9 & 51.4 $\rightarrow$ 26.6 \\
				& 0.5 & 48.5 $\rightarrow$ 6.50 & 51.4 $\rightarrow$ 11.0 \\
				
				\bottomrule
			\end{tabularx}
		\end{center}
		\vspace{-0.3in}
	\end{table}
	
	\section{Methodology}
	We first briefly recall the preliminary of UDA, adversarial training, and self-supervision. 
	We then perform the first-of-its-kind empirical study to show that existing UDA methods are vulnerable to adversarial attacks, which arises tremendous concerns for the application of these methods in safety-critical areas.
	To address this problem, we propose a new domain adaptation method known as ASSUDA to improve the model robustness without satisfying the predictive accuracy.
	Specifically, our method takes advantage of adversarial training and self-supervision and thus enabling us to generate more robust and generalized features.

	\subsection{Preliminary}
	\paragraph{UDA in Semantic Segmentation}
	Consider the problem of UDA in semantic segmentation, we have the source-domain data $ \{\mathcal{X}_s, Y_s\} $ and the target-domain data $ \mathcal{X}_t $. 
	Our goal is to learn a segmentation model which guarantees accurate prediction on the target domain.
	Formally, the loss function of a typical UDA model is given by,
	\begin{equation} \label{eq:1}
	\begin{aligned}
	\mathcal{L}_{seg}(\mathcal{X}_s, Y_s; \theta_C) + 
	\alpha \mathcal{L}_{dis}(\mathcal{X}_s, \mathcal{X}_{t}),
	\end{aligned}
	\end{equation}
	where $ \mathcal{L}_{seg} $ is the typical segmentation objective parameterized by $\theta_C$, and $\mathcal{L}_{dis}$ measures the domain distance.
	The most commonly used $\mathcal{L}_{dis}$ is the adversarial loss that encourages a discriminative and domain-invariant feature representation through a domain discriminator $D(\cdot)$ \cite{hoffman2016fcns,hoffman2017cycada,tsai2018learning}, which is formalized as $\mathcal{L}_{dis}(\mathcal{X}_s, \mathcal{X}_{t}; \theta_D)$.

	\vspace{-10pt}
	\paragraph{Adversarial Training}
	Recall that the objective of an vanilla adversarial training is:
	\begin{equation} \label{eq:2}
	\begin{aligned}
	\underset{x}{\arg\min} \hspace{2pt} \mathbb{E}_{(x, y) \sim \mathbb{D}} [\underset{\epsilon \in \mathbb{S}}{\max} \hspace{2pt} \mathcal{L}(f_{\theta}(x + \eta), y)] 
	\end{aligned}
	\end{equation}
	where $\mathbb{S}$ are allowed perturbations, $\tilde{x} \leftarrow x + \eta$ is an adversarial example of $x$ with the perturbation $\eta$. 
	To obtain $\eta$, the most commonly used attack method is FGSM \cite{goodfellow2014explaining}:
	\begin{equation} \label{eq:3}
	\begin{aligned}
	\eta = \epsilon \text{sign} (\bigtriangledown_x \mathcal{L}(f_{\theta}(x), y)),
	\end{aligned}
	\end{equation}
	where $\epsilon$ is the magnitude of the perturbation.
	The generated adversarial examples $\tilde{x}$ are imperceptible to human but can easily fool deep neural networks.
	Recent studies further prove that training models exclusively on adversarial examples can improve the model robustness \cite{madry2017towards}.

	\begin{figure}
		\begin{center}
			\includegraphics[width=1\linewidth]{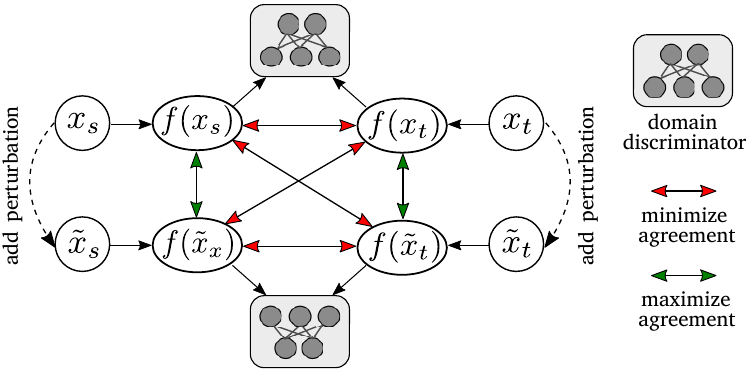}
		\end{center}
		\caption{An overview of the proposed method. For each sampled pair of source image $x_s$ and target image $x_t$, we generate their adversarial example $\tilde{x}_s$ and $\tilde{x}_t$, respectively. A segmentation model $f(\cdot)$ and a domain discriminator are trained to maximize/minimize agreement and align cross-domain representations.}
		\label{fig:framework}
		\vspace{-0.2in}
	\end{figure}
	
	\subsection{Robustness of UDA Methods}
	Although existing UDA methods achieve record-breaking predictive accuracy, their robustness against adversarial attacks remains unexplored.
	We hypothesis that they are also vulnerable to adversarial attacks, which makes it risky to apply those UDA methods in safety-critical environments.
	To fill this gap and to validate our hypothesis, we perform the black-box attack on BDL \cite{li2019bidirectional} by conducting the following two steps:
	(1) for each clean image in the test data, we first generate its adversarial example by attacking PSPNet \cite{zhao2017pyramid} with $\epsilon=0.1$, $\epsilon=0.25$ and $\epsilon=0.5$, respectively; 
	(2) we then evaluate the pre-trained BDL model on the generated adversarial examples (or perturbed test data) (Figure~\ref{fig:attack_BDL_fig}).
	The rationale behind this setting is that 
	(i) most of existing state-of-the-art UDA methods in semantic segmentation \cite{huang2020contextual,pan2020unsupervised,wang2020differential,kim2020learning,yang2020fda,yang2020label,pan2020two} are built upon BDL, so conducting empirical studies on this method would be representative;
	(ii) a black-box attack assumes that the attacker can only access very limited information of the victim model, which is a common case in the real world.
	The commonly used attack strategy is: attackers first train a surrogate model to approximate the information of the victim model and then generate adversarial examples to attack the victim model.
	This strategy is motivated by the fact that adversarial attacks are transferable across different models \cite{goodfellow2014explaining}, \ie, the adversarial examples generated to attack a specific model are also harmful to other models.
	Therefore, a black-box attack would be very dangerous if it can work.
	We hereby perform the black-box attack to examine the transferability of adversarial examples on UDA models.
	
	As shown in Table~\ref{table:attack_BDL}, despite the remarkable performance of BDL on clean test data, slight and unnoticeable perturbations can result in dramatic performance degradation.
	For instance, BDL (with VGG16 backbone) only achieves a mean IoU (mIoU) of 30.5\% on perturbed test data generated by $\epsilon=0.1$, compared to 41.3\% on clean images.
	By increasing the perturbation ratio $\epsilon$, the performance can drop even further (Figure~\ref{fig:attack_BDL_fig}), indicating that BDL can be easily fooled by slight perturbations on the test data, even though the perturbation is generated by a surrogate model.
	This empirical study suggests that existing UDA methods are also possibly vulnerable to adversarial perturbations, which can make them especially risky for some security-related areas.
	
	\subsection{Adversarial Self-Supervision UDA}
	To address this problem, the most straightforward approach is adversarial training (equation \ref{eq:2}) which requires class labels to generate adversarial examples. 
	However, we are unable to access the labels of target data under the scenario of UDA (equation \ref{eq:1}).
	The success of existing UDA methods heavily relies on the self-training strategy that alternatively generates highly confident pseudo labels for the target domain and re-trains the model using these labels \cite{li2019bidirectional,huang2020contextual,pan2020unsupervised,wang2020differential,kim2020learning,yang2020fda,yang2020label,pan2020two}.
	Although pseudo labels provide an opportunity to generate adversarial examples for the target data, these labels are usually noisy and less accurate.
	Hendrycks \etal prove that self-supervision improves the robustness of deep neural networks for vision tasks \cite{hendrycks2019using}.
	Nevertheless, commonly used pretext tasks (\eg, rotation prediction and jigsaw) in self-supervision fail to provide the critical supervision signals in learning discriminative features for semantic segmentation.
	
	These challenges reach the question: can we take advantage of both adversarial training and self-supervision in improving the robustness of UDA methods in semantic segmentation?
	To achieve this goal, we propose to build the pretext task by using adversarial examples (Figure~\ref{fig:framework}).
	Specifically, we consider a clean image and its adversarial example as a positive pair and maximize agreement on their segmentation outputs by a contrastive loss. 
	This is motivated by the fact that a clean image and its adversarial example should share the same segmentation output.
	Different from \cite{chen2020simple} (designed for image classification) that uses a contrastive loss in the latent space, our pretext task is performed in the output space to learn discriminative representations for semantic segmentation. 
	To adapt knowledge from the source domain to the target domain, a domain discriminator is applied to the source and target outputs.
	It is worth mentioning that the domain discriminator minimizes the domain-level difference, while the contrastive loss is performed on the pixel level. 
	
	
	Our model is built upon BDL \cite{li2019bidirectional}, and follows the same experimental protocol to generate the transformed source images $\mathcal{X}_{s \rightarrow t}$ and pseudo labels $Y_{t'}$ of $\mathcal{X}_{t}$.
	For simplicity, we use $\mathcal{X}_{s}$ to represent $\mathcal{X}_{s \rightarrow t}$ in the remaining of this paper, unless otherwise specified.
	At each training iteration $t$, a minibatch of $N$ source-target pairs are randomly sampled from $\mathcal{X}_{s}$ and $\mathcal{X}_{t}$, resulting in $2N$ examples: $\{x_s^{(i)}, x_t^{(i)}\}_{i=1}^N$.
	Their adversarial examples $\{\tilde{x}_s^{(i)}, \tilde{x}_t^{(i)}\}_{i=1}^N$ are generated by fixing $\theta_C$ and $\theta_D$ that are learned from the previous iteration $t-1$,
	\begin{equation} \label{eq:4}
	\begin{aligned}
	\tilde{{x}}_{s}^{(i)} = {} & 
	{x}_{s}^{(i)} + \epsilon_m \text{sign} (\bigtriangledown_x [\mathcal{L}_{seg}({x}_s^{(i)}, y_s^{(i)}; \theta_C) + \mathcal{L}_{adv}(x_s^{(i)}; \theta_D)])
	\end{aligned}
	\end{equation}
	\begin{equation} \label{eq:5}
	\begin{aligned}
	\tilde{{x}}_{t}^{(i)} = {} & 
	{x}_{t}^{(i)} + \epsilon_m \text{sign} (\bigtriangledown_x [\mathcal{L}_{seg}({x}_t^{(i)}, y_{t'}^{(i)}; \theta_C) + \mathcal{L}_{adv}(x_t^{(i)}; \theta_D)])
	\end{aligned}
	\end{equation}
	where $D(\cdot)$ is a domain discriminator parameterized by $\theta_D$, $\mathcal{L}_{adv}$ is an adversarial loss which is designed for encouraging cross-domain knowledge alignment:
	\begin{equation}
	\begin{aligned}
	\mathcal{L}_{adv}(x; \theta_D) = {} & 
	\mathop{\mathbb{E}}[log{D(f_{\theta_C}(x_s))}] + \\ &
	\mathop{\mathbb{E}}[log{(1 - D(f_{\theta_C}(x_{t})))}]
	\end{aligned}
	\end{equation}
	
	Given these $4N$ data points $\{x_s^{(i)}, x_t^{(i)}, \tilde{x}_s^{(i)}, \tilde{x}_t^{(i)}\}_{i=1}^N$, each pair of examples $\{x_\alpha^{(i)}, \tilde{x}_\alpha^{(i)} \}$ is considered as a positive pair ($\alpha$ can be either $s$ or $t$ to denote a source or a target domain), while the other $4N-2$ examples are considered as negative examples.
	We define the contrastive loss for a positive pair $(i, j)$ as
	\begin{equation}
	\begin{aligned}
	\ell_{i,j} = {} & 
	-log \frac{exp(sim(f_{\theta_C}(x^{(i)}), f_{\theta_C}(x^{(j)})))}{\sum_{k=1}^{4N} \mathbbm{1}_{[k \ne i]} exp(sim(f_{\theta_C}(x^{(i)}), f_{\theta_C}(x^{(k)})))},
	\end{aligned}
	\end{equation}
	where $sim(\mathbf{U}, \mathbf{V}) = exp(-dist(\mathbf{U}, \mathbf{V})/(2 \sigma^2))$ is Gaussian kernel that is used to measure the similarity between two segmentation output tensors $\mathbf{U}$ and $\mathbf{V}$, $dist(;)$ is the Euclidean distance. 
	The contrastive loss $\mathcal{L}_{con}({x}_s, \tilde{x}_s, {x}_t, \tilde{x}_t; \theta_C, \theta_D))$ is computed across all positive pairs (see Algorithm~\ref{alg:alg_1}).
	
	Taken together, our goal is to minimize the following loss function:
	\begin{equation} \label{eq:8}
	\begin{aligned}
	\mathcal{L}_{total} = {} & 
	\mathcal{L}_{seg}({x}_s, y_s; \theta_C) + \mathcal{L}_{seg}(\tilde{x}_s, y_s; \theta_C) + \\ &
	\mathcal{L}_{seg}({x}_t, y_{t'}; \theta_C) + \mathcal{L}_{seg}(\tilde{x}_t, y_{t'}; \theta_C) + \\ &
	\delta(\mathcal{L}_{con}({x}_s, \tilde{x}_s, {x}_t, \tilde{x}_t; \theta_C, \theta_D)) + \\ &
	\gamma(\mathcal{L}_{adv}(x; \theta_D))
	\end{aligned}
	\end{equation}
	Therefore, our model can leverage the regularization power of adversarial examples through a self-supervision manner, and in turn, improve the model robustness against adversarial attacks.
	The whole training process is detailed in Algorithm~\ref{alg:alg_1}.
	
	\setlength{\textfloatsep}{10pt}%
	\begin{algorithm}[t]
		\SetAlgoLined
		\footnotesize
		\textbf{Input:} Source data $ \{\mathcal{X}_s, Y_s\} $ and target data $ \{\mathcal{X}_t\} $, \\
		\hspace{23pt} segmentation model initialized as $\theta_C^0$, \\
		\hspace{23pt} domain discriminator initialized as $\theta_D^0$, \\
		\hspace{23pt} batch size $N$, number of training iteration $T$ \\
		\KwResult{$\theta_{C}^T$ and $\theta_{D}^T$}
		\For{$t\leftarrow 1$ \KwTo $T$}{
			Sample a batch of source-target pairs $\{x_s^{(k)}, x_t^{(k)}\}_{k=1}^N$  \\
			\# \textit{adversarial attack}
			
			\For {$k\in \{1, ..., N\}$}{
				Generate adversarial examples: $\{\tilde{x}_s^{(k)}, \tilde{x}_t^{(k)}\}_{k=1}^N$ \\
				Define $ x^{(4k-3)} = x_s^{(k)}, x^{(4k-2)} = x_t^{(k)}, x^{(4k-1)} = \tilde{x}_s^{(k)}, x^{(4k)} = \tilde{x}_t^{(k)} $
			}
			
			\# \textit{adversarial self-supervision}
			
			\For {$i\in \{1, ..., 4N\}$ and $j\in \{1, ..., 4N\}$}{
				$s_{i,j} = exp(\frac{-dist\Big(f_{\theta_C^{t-1}}(x^{(i)}), f_{\theta_C^{t-1}}(x^{(j)})\Big)}{(2 \sigma^2)})$
			}
			
			Define $\ell_{i,j} =
			-log \frac{exp(s_{i,j})}{\sum_{k=1}^{4N} \mathbbm{1}_{[k \ne i]} exp(s_{i,k})}$
			
			\# \textit{contrastive loss}
			
			$\mathcal{L}_{con} = \frac{1}{4N} \sum_{k=1}^{N}[\ell_{4k-3,4k-1} + \ell_{4k-1,4k-3} + \ell_{4k-2,4k} + \ell_{4k,4k-2}] $
			
			\# \textit{update model parameters}
			
			$\theta_C^t \leftarrow \theta_C^{t-1} - \beta \bigtriangledown_{\theta_C} \mathcal{L}_{total} $ 
			
			$\theta_D^t \leftarrow \theta_D^{t-1} - \beta \bigtriangledown_{\theta_D} \mathcal{L}_{total}  $
		}
		\textbf{return} $\theta_{C}^T$ and $\theta_{D}^T$
		
		\caption{The whole training process.}
		\label{alg:alg_1}
	\end{algorithm}

	\begin{table*}[t]
		\caption{Quantitative study of "GTA5 to Cityscapes". VGG16 (upper part) and ResNet101 (lower part) are used as backbones in this experiment. The performance is measured on 19 common classes with criteria: per-class IoU, mean IoU (mIoU), and mIoU drop (performance degradation of the model after being attacked). The higher the mIoU and the lower the mIoU drop, the more robust the model. The best result in each column is highlighted in bold.}
		\label{table:gta2city}
		
		\footnotesize
		\setlength\tabcolsep{2.5pt}
		\begin{center}
			\begin{tabular}{ @{} l|c|*{19}{c}|*{2}{c} @{} }
				\toprule
				\multicolumn{23}{ c }{\bf GTA5 to Cityscapes } \\
				\midrule
				& $\mathbf{\epsilon}$ & \rotatebox[origin=c]{90}{road} & \rotatebox[origin=c]{90}{sidewalk} & \rotatebox[origin=c]{90}{building} & \rotatebox[origin=c]{90}{wall} & \rotatebox[origin=c]{90}{fence} & \rotatebox[origin=c]{90}{pole} & \rotatebox[origin=c]{90}{  traffic light  } & \rotatebox[origin=c]{90}{traffic sign} & \rotatebox[origin=c]{90}{vegetation} & \rotatebox[origin=c]{90}{terrain} & \rotatebox[origin=c]{90}{sky} & \rotatebox[origin=c]{90}{person} & \rotatebox[origin=c]{90}{rider} & \rotatebox[origin=c]{90}{car} & \rotatebox[origin=c]{90}{truck} & \rotatebox[origin=c]{90}{bus} & \rotatebox[origin=c]{90}{train} & \rotatebox[origin=c]{90}{motorbike} & \rotatebox[origin=c]{90}{bicycle} & \rotatebox[origin=c]{90}{\bf mIoU } &
				\rotatebox[origin=c]{90}{\bf mIoU drop} \\
				\midrule
				
				FDA \cite{yang2020fda} & \multirow{5}{*}{$0.1$} &
				73.9 & 18.5 & 69.7 & 7.5 & 6.4 & \bf18.7 & 23.9 & 21.5 & 76.7 & 12.2 & 66.3 & 45.2 & 18.4 & 70.2 & 18.9 & 13.9 & \bf14.6 & 9.3 & 22.0 & 32.0 & 10.2\\ 
				
				AdaptSegNet \cite{tsai2018learning} & &
				71.9 & 22.7 & 70.8 & 7.6 & 7.9 & 16.5 & 15.4 & 8.3 & 71.8 & 12.2 & 52.6 & 33.8 & 0.6 & 65.8 & 15.8 & 7.6 & 0.0 & 0.7 & 0.1 & 25.4 & 9.6\\ 
				
				PCEDA \cite{yang2020phase} & & 
				\bf90.9 & 25.0 & 73.5 & 6.3 & 7.2 & 14.2 & \bf24.0 & \bf27.4 & 76.2 & 23.4 & 
				70.3 & 45.0 & 19.9 & 70.0 & 16.3& \bf20.3 & 0.0 & 9.8 & \bf25.1 & 33.4 &11.2\\
				
				BDL \cite{li2019bidirectional} & & 
				64.0 & 21.9 & 70.0 & 10.0 & 3.9 & 8.4 & 20.5 & 12.8 & 77.4 & 22.3 & 
				79.2 & 49.8 & 13.8 & 73.2 & 17.8& 12.1 & 0.0 & 7.8 & 15.2 & 30.5 & 10.8\\
				
				\rowcolor{LightCyan}
				Ours & &  
				90.7 & \bf40.9 & \bf80.3 & \bf24.6 & \bf14.9 & 13.1 & 22.8 & 16.7 & \bf83.0 & \bf36.1 & \bf81.7 & \bf52.7 & \bf24.3 & \bf82.1 & \bf20.8 & 19.1 & 1.8 & \bf19.5 & 23.6 & \bf39.4 & \bf0.7\\
				\midrule
				
				FDA & \multirow{5}{*}{$0.25$} &
				25.4 & 3.4 & 24.5 & 0.5 & 1.6 & 2.4 & 7.7 & 6.4 & 58.6 & 1.2 & 
				44.8 & 6.5 & 1.4 & 14.6 & 4.9 & 0.4 & \bf0.1 & 0.1 & 1.3 & 10.8 & 31.4\\ 
				
				AdaptSegNet & & 
				5.4 & 5.0 & 43.8 & 1.2 & 2.2 & 3.7 & 6.3 & 2.5 & 31.3 & 3.9 & 22.8 & 6.2 & 0.0 & 11.9 & 4.3 & 0.1 & 0.0 & 0.0 & 0.0 & 7.9 & 27.1\\
				
				PCEDA & & 
				34.6 & 1.5 & 40.9 & 0.6 & 1.6 & 2.2 & 9.6 & 11.1 & 56.4 & 0.5 & 
				43.8 & 12.7 & 2.0 & 28.0 & 7.0& 3.7 & 0.0 & 1.0 & 5.0 & 13.8 &30.8 \\
				
				BDL & & 
				25.4 & 4.7 & 55.1 & 2.8 & 1.5 & 1.3 & 9.1 & 4.3 & 61.3 & 1.5 & 
				54.1 & 26.7 & 0.1 & 20.7 & 6.5& 1.5 & 0.0 & 0.7 & 1.0 & 14.6 & 26.7\\
				
				\rowcolor{LightCyan}
				Ours & &  
				\bf88.5 & \bf20.6 & \bf77.3 & \bf8.4 & \bf10.8 & \bf7.2 & \bf19.2 & \bf14.9 & \bf77.8 & \bf22.6 & \bf84.6 & \bf48.4 & \bf18.2 & \bf75.6 & \bf12.1 & \bf13.9 & 0.0 & \bf6.1 & \bf17.4 & \bf32.8 & \bf7.3\\
				\midrule
				
				FDA & \multirow{5}{*}{$0.5$} &
				22.0 & 0.4 & 3.2 & 0.0 & 1.3 & 0.1 & 1.9 & 0.6 & 33.8 & 1.1 & 
				22.6 & 0.1 & 0.0 & 0.1 & 0.0 & 0.0 & 0.0 & 0.0 & 0.0 & 4.6 & 37.6 \\ 
				
				AdaptSegNet & & 
				0.1 & 0.0 & 14.4 & 0.0 & 2.1 & 0.7 & 2.9 & 0.4 & 23.3 & 0.0 & 8.4 & 0.2 & 0.0 & 0.1 & 0.0 & 0.0 & 0.0 & 0.0 & 0.0 & 2.8 & 32.2\\
				
				PCEDA & & 
				26.8 & 0.1 & 15.0 & 0.1 & 1.3 & 0.1 & 2.5 & 2.3 & 18.1 & 0.0 & 
				15.4 & 0.1 & 0.0 & 2.0 & 0.2& 0.0 & 0.0 & 0.0 & 0.0 & 4.4 &40.2\\
				
				BDL & & 
				27.8 & 0.9 & 36.8 & 0.5 & 1.2 & 0.1 & 2.7 & 0.9 & 34.1 & 0.0 & 
				25.1 & 5.4 & 0.0 & 0.7 & 0.0& 0.0 & 0.0 & 0.0 & 0.0 & 7.1 & 34.2 \\
				
				\rowcolor{LightCyan}
				Ours & &  
				\bf54.0 & \bf2.1 & \bf66.2 & \bf0.9 & \bf3.3 & \bf1.0 & \bf13.0 & \bf8.8 & \bf62.0 & \bf3.9 & \bf73.8 & \bf29.1 & \bf1.4 & \bf35.3 & \bf3.2 & \bf2.5 & 0.0 & 0.0 & \bf2.9 & \bf19.1 & \bf21.0\\
				
				\midrule                    
				\midrule
				
				FDA \cite{yang2020fda} & \multirow{9}{*}{$0.1$} &
				85.8 & 27.8 & 70.2 & 8.6 & 7.4 & 17.9 & 30.7 & 23.4 & 70.8 & 22.4 & 59.7 & 53.8 & 26.5 & 71.6 & 29.2 & 26.8 & \bf6.3 & 23.1 & \bf38.3 & 36.9 & 13.5\\ 
				
				FADA \cite{wang2020classes} & &
				53.2 & 19.7 & 65.2 & 6.3 & 14.1 & 21.3 & 19.0 & 8.2 & 74.4 & 21.6 & 55.7 & 50.3 & 14.8 & 73.2 & 13.4 & 9.1 & 1.0 & 9.6 & 20.5 & 29.0 & 20.2 \\
				
				IntraDA \cite{pan2020unsupervised} & &
				89.1 & 31.1 & 76.6 & 11.3 & 16.4 & 14.9 & 25.3 & 15.8 & 80.8 & 29.4 & 74.9 & 54.3 & 23.3 & 78.7 & 32.1 & \bf39.2 & 0.0 & 21.5 & 30.8 & 39.2 & 7.1\\
				
				CLAN \cite{luo2019taking} & & 
				75.8 & 21.3 & 69.8 & 11.9 & 7.3 & 12.7 & 24.6 & 8.8 & 77.1 & 20.4 & 66.9 & 51.0 & 19.6 & 65.4 & 28.7 & 31.3 & 2.5 & 15.2 & 24.8 & 33.4 & 9.8 \\
				
				MaxSquare \cite{maxsquareloss} & & 
				28.6 & 9.3 & 52.0 & 3.9 & 3.1 & 9.7 & 29.1 & 10.3 & 73.6 & 10.2 & 41.7 & 46.1 & 19.1 & 36.1 & 26.5 & 10.7 & 0.2 & 17.2 & 28.0 & 24.0 & 22.4 \\
				
				AdaptSegNet \cite{tsai2018learning} & & 
				80.9 & 21.2 & 66.3 & 7.4 & 5.7 & 7.4 & 25.2 & 6.5 & 76.2 & 12.5 & 69.9 & 45.6 & 11.7 & 71.3 & 21.8 & 8.0 & 1.6 & 6.5 & 14.3 & 29.5 & 12.9 \\
				
				PCEDA \cite{yang2020phase} & & 
				\bf89.8 & 31.8 & 75.8 & 17.4 & 9.2 & 26.9 & \bf31.1 & 30.0 & 80.0 & 19.3 & \bf85.6 & \bf55.2 & 27.5 & 79.4 & 30.2 & 34.4 & 0.0 & 20.3 & \bf38.3 & 41.2 & 9.3\\
				
				BDL \cite{li2019bidirectional} & & 
				75.5 & 31.3 & 75.3 & 8.8 & 8.5 & 17.1 & 29.3 & 23.0 & 76.9 & 22.4 &
				80.5 & 51.2 & 25.8 & 51.9 & 24.0& 33.3 & 1.6 & 20.3 & 31.3 & 36.2 & 12.3\\
				
				\rowcolor{LightCyan}
				Ours & & 
				88.8 & \bf41.3 & \bf81.8 & \bf26.0 & \bf17.4 & \bf27.8 & 30.2 & \bf37.9 & \bf81.6 & \bf30.5 & 81.2 & 55.0 & \bf29.4 & \bf79.7 & \bf33.0 & 38.4 & 0.0 & \bf32.7 & 34.8 & \bf44.6 & \bf1.8\\
				\midrule
				
				FDA  & \multirow{9}{*}{$0.25$} &
				50.8 & 6.7 & 51.0 & 1.6 & 3.7 & 3.5 & 17.2 & 6.3 & 49.5 & 1.5 & 60.9 & 28.3 & 12.8 & 49.1 & 14.5 & 4.6 & 1.2 & 2.6 & 25.0 & 20.6 & 29.8 \\ 
				
				FADA & & 
				54.1 & 14.8 & 50.4 & 2.2 & \bf8.2 & 6.8 & 4.7 & 0.9 & 59.4 & 7.4 & 32.8 & 29.9 & 3.0 & 53.6 & 4.1 & 0.3 & 1.2 & 0.7 & 5.9 & 17.9 & 31.3\\ 
				
				IntraDA & & 
				26.4 & 3.0 & 46.3 & 0.4 & 4.5 & 0.7 & 8.6 & 0.5 & 30.9 & 0.4 & 43.9 & 21.3 & 1.2 & 47.5 & 8.33 & 7.5 & 0.0 & 0.2 & 6.5 & 13.6 & 32.7 \\
				
				CLAN & & 
				58.3 & 9.4 & 52.7 & 5.0 & 2.7 & 1.3 & 14.7 & 2.1 & 58.5 & 3.0 & 64.5 & 37.6 & 14.0 & 46.1 & \bf20.0 & 13.6 & \bf1.8 & 3.6 & 17.3 & 22.4 & 20.8 \\
				
				MaxSquare & & 
				15.2 & 2.3 & 37.9 & 2.7 & 1.5 & 1.0 & 15.8 & 1.8 & 54.1 & 1.5 & 30.6 & 14.3 & 7.2 & 31.5 & 11.8 & 1.6 & 0.0 & 0.7 & 13.8 & 12.9 & 33.5 \\
				
				AdaptSegNet & & 
				66.9 & 4.8 & 32.8 & 1.3 & 2.4 & 0.7 & 13.2 & 1.2 & 60.6 & 2.4 & 65.3 & 19.6 & 1.5 & 49.0 & 8.2 & 1.2 & 0.0 & 0.1 & 0.8 & 17.5 & 24.9 \\
				
				PCEDA & & 
				\bf76.4 & 3.0 & 50.9 & 1.5 & 3.3 & 11.5 & 18.1 & 10.0 & 59.3 & 0.6 & 59.4 & 37.0 & 16.1 & 49.6 & 11.6& 5.6 & 0.0 & 2.6 & \bf25.2 & 23.3 & 27.2 \\
				
				BDL & & 
				40.7 & 7.2 & 56.6 & 3.1 & 2.0 & 4.0 & 20.3 & 5.5 & 62.7 & 1.5 &
				65.8 & 19.4 & 15.3 & 30.2 & 8.0& 8.4 & 0.0 & 6.4 & 21.2 & 19.9 & 28.6\\
				
				\rowcolor{LightCyan}
				Ours & & 
				74.1 & \bf26.3 & \bf69.9 & \bf9.1 & 5.6 & \bf20.7 & \bf25.0 & \bf29.6 & \bf68.0 & \bf11.7 &
				\bf77.5 & \bf43.0 & \bf20.2 & \bf66.8 & 19.8 & \bf24.9 & 0.0 & \bf12.6 & 25.0 & \bf33.1 & \bf13.3\\
				\midrule
				
				FDA & \multirow{9}{*}{$0.5$} &
				14.5 & 0.9 & 23.2 & 1.0 & \bf5.3 & 1.1 & 7.6 & 0.9 & 28.4 & 0.0 & 57.9 & 3.0 & 0.2 & 8.2 & 3.8 & 0.0 & 0.0 & 0.0 & 1.6 & 8.3 & 42.1 \\ 
				
				FADA & & 
				17.4 & 7.6 & 18.1 & 1.2 & 2.1 & 0.4 & 0.5 & 0.1 & 29.2 & 0.0 & 11.8 & 3.8 & 0.2 & 18.5 & 0.0 & 0.1 & 0.0 & 0.0 & 0.0 & 5.8 & 43.4 \\ 
				
				IntraDA & & 
				26.4 & 3.0 & 46.3 & 0.4 & 4.5 & 0.7 & 8.6 & 0.5 & 30.9 & 0.4 & 43.9 & 21.3 & 1.2 & \bf47.5 & 8.3 & \bf7.5 & 0.0 & 0.2 & 6.5 & 13.6 & 32.7 \\
				
				CLAN & & 
				33.0 & 0.6 & 39.2 & \bf2.3 & 1.8 & 0.1 & 8.4 & 0.2 & 36.2 & 0.3 & 38.1 & \bf21.5 & 3.4 & 38.0 & 9.4 & 3.4 & 0.0 & 0.1 & 4.3 & 12.6 & 30.6 \\
				
				MaxSquare & & 
				17.0 & 0.3 & 33.6 & 0.6 & 2.2 & 0.4 & 9.9 & 0.4 & 29.5 & 0.0 & 31.2 & 3.5 & 0.4 & 28.8 & 5.7 & 0.4 & 0.0 & 0.0 & 1.3 & 8.7 & 37.7 \\
				
				AdaptSegNet & & 
				\bf43.0 & 0.2 & 10.1 & 0.7 & 2.8 & 0.2 & 7.3 & 0.1 & 34.8 & 0.0 & 58.1 & 4.9 & 0.0 & 18.6 & 0.8 & 0.3 & 0.0 & 0.0 & 0.0 & 9.6 & 32.8 \\
				
				PCEDA & & 
				30.4 & 0.0 & 36.6 & 0.2 & 1.7 & 1.5 & 4.0 & 1.2 & 27.1 & 0.0 & 
				8.1 & 9.7 & 0.4 & 7.4 & 1.2& 0.0 & 0.0 & 0.0 & 5.3 & 7.1 & 43.4\\
				
				BDL & & 
				9.7 & 0.1 & 25.9 & 0.0 & 0.8 & 0.2 & 8.1 & 0.6 & 43.5 & 0.0 &
				13.7 & 4.8 & 4.3 & 7.6 & 2.6& 0.0 & 0.0 & 0.2 & 1.9 & 6.5 & 42.0\\
				
				\rowcolor{LightCyan}
				Ours & &  
				40.8 & \bf11.0 & \bf47.5 & 2.1 & 1.9 & \bf10.2 & \bf16.5 & \bf13.6 & \bf45.5 & \bf0.5 & \bf72.6 & 14.4 & \bf6.0 & 23.4 & \bf13.4 & 4.2 & 0.0 & \bf0.7 & \bf9.4 & \bf17.6 & \bf28.8\\
				
				\bottomrule
			\end{tabular}
		\end{center}
		\vspace{-0.3in}
	\end{table*}

	\newcommand{\xmark}{\ding{55}}%
	\begin{table*}[t]
		\caption{Quantitative study of "SYNTHIA to Cityscapes". VGG16 (upper part) and ResNet101 (lower part) are used as backbones in this experiment. The comparison is performed on 16 common classes for VGG16 and 13 common classes for ResNet101.}
		
		\label{table:synthia2city}
		\footnotesize
		\setlength\tabcolsep{2.5pt}
		\begin{center}
			\begin{tabular}{ @{} l|c|*{16}{c}|*{2}{c} @{} }
				\toprule
				\multicolumn{20}{ c }{\bf SYNTHIA to Cityscapes } \\
				\midrule\            
				
				& $\mathbf{\epsilon}$ & \rotatebox[origin=c]{90}{road} & \rotatebox[origin=c]{90}{sidewalk} & \rotatebox[origin=c]{90}{building} & \rotatebox[origin=c]{90}{wall} & \rotatebox[origin=c]{90}{fence} &\rotatebox[origin=c]{90}{pole} & \rotatebox[origin=c]{90}{traffic light} & \rotatebox[origin=c]{90}{traffic sign} & \rotatebox[origin=c]{90}{vegetation} & \rotatebox[origin=c]{90}{sky} & \rotatebox[origin=c]{90}{person} & \rotatebox[origin=c]{90}{rider} & \rotatebox[origin=c]{90}{car} & \rotatebox[origin=c]{90}{bus} & \rotatebox[origin=c]{90}{motorbike} & \rotatebox[origin=c]{90}{bicycle} & 
				\rotatebox[origin=c]{90}{\bf mIoU } &
				\rotatebox[origin=c]{90}{\bf mIoU drop} \\ 
				\midrule
				
				FDA \cite{yang2020fda} & \multirow{4}{*}{$0.1$} &
				68.5 & 28.4 & 72.7 & 0.4 & 0.3 & 22.2 & 5.1 & 19.1 & 57.6 & 75.7 & 45.8 & 18.8 & 55.6 & 18.5 & 5.1 & 31.5 & 32.8 & 7.7 \\
				
				PCEDA \cite{yang2020phase} & & 
				80.9 & 25.0 & 73.5 & \bf6.3 & \bf7.1 & 14.2 & \bf24.0 & 27.4 & 76.2 & 
				70.3 & 45.0 &19.9 & 70.0& \bf20.3 & \bf9.8 &25.1&37.2 &3.9\\
				
				BDL \cite{li2019bidirectional} & & 
				34.9 & 21.2 & 47.8 & 0.0 & 0.2 & 20.5 & 9.2 & 20.2 & 67.2 &
				74.3 & 49.0 & 17.5 & 57.2 & 11.9 & 2.5 & 34.6 & 29.3 & 9.7\\
				
				\rowcolor{LightCyan}
				Ours & &  
				\bf88.4 & \bf43.4 & \bf77.7 & 0.1 & 0.2 & \bf25.6 & 10.9 & \bf27.8 & \bf78.7 & \bf80.1 & \bf55.3 & \bf21.8 & \bf76.5 & 18.2 & 6.8 & \bf42.9 & \bf40.9 & \bf-0.8\\
				\midrule
				
				FDA & \multirow{4}{*}{$0.25$} &
				46.3 & 16.0 & 38.7 & 0.0 & 0.2 & 4.9 & 2.5 & 8.9 & 31.3 & 38.9 & 8.6 & 5.3 & 17.7 & 6.0 & 1.3 & 5.4 & 14.5 & 26.0 \\
				
				PCEDA & & 
				75.6 & 11.4 & 59.1 & 0.0 & \bf0.4 & 9.6 & 5.5 & 12.9 & 63.1 & 45.0 & 
				30.7 & 13.4 & 34.9 & 8.6 & 2.5& 24.5& 24.8 & 16.3\\
				
				BDL & & 
				8.0 & 8.9 & 31.1 & 0.0 & 0.14 & 8.7 & 6.9 & 9.8 & 52.0 &
				54.1 & 22.9 & 4.9 & 25.6 & 2.5 & 0.8 & 13.3 & 13.6 & 25.4\\
				
				\rowcolor{LightCyan}
				Ours & &  
				\bf84.0 & \bf29.7 & \bf69.1 & 0.0 & 0.2 & \bf22.1 & \bf10.2 & \bf24.9 & \bf70.1 & \bf56.7 & \bf45.9 & \bf17.2 & \bf59.2 & \bf13.7 & \bf3.2 & \bf31.5 & \bf33.6 & \bf6.5\\
				\midrule
				
				FDA & \multirow{4}{*}{$0.5$} & 
				42.2 & 4.9 & 14.2 & 0.0 & 0.1 & 0.6 & 1.0 & 1.7 & 26.2 & 1.9 & 0.5 & 0.4 & 1.5 & 0.1 & 0.1 & 0.1 & 6.0 & 34.5 \\
				
				PCEDA & & 
				\bf66.2 & 1.1 & \bf47.9 & 0.0 & \bf0.4 & 3.1 & 2.5 & 5.0 & 47.8 & \bf18.8 & 10.0 & 1.9 & 8.3 & 3.2 & 1.1 & 10.2 & 14.2 &26.9 \\
				
				BDL & & 
				0.6 & 1.0 & 24.8 & 0.0 & 0.0 & 1.6 & 1.9 & 2.3 & 35.8 & 18.6 & 2.2 & 0.1 & 4.1 & 0.1 & 0.0 & 0.5 & 5.9 & 33.1\\
				
				\rowcolor{LightCyan}
				Ours & &  
				65.4 & \bf9.3 & 46.9 & 0.0 & \bf0.4 & \bf15.0 & \bf7.0 & \bf14.5 & \bf48.6 & 9.9 & \bf22.8 & \bf7.6 & \bf26.9 & \bf4.4 & \bf1.5 & \bf15.6 & \bf18.5 & \bf21.6\\
				
				\midrule
				\midrule
				
				FDA \cite{yang2020fda} & \multirow{8}{*}{$0.1$} &
				83.4 & 32.4 & 73.5 & \xmark & \xmark & \xmark & 13.1 & \bf18.9 & \bf71.6 & 79.5 & \bf56.1 & 24.9 & 77.5 & 27.6 & \bf18.2 & \bf42.8 & 47.7 & 4.8 \\
				
				FADA \cite{wang2020classes} & &
				74.0 & 32.5 & 69.8 & \xmark & \xmark & \xmark & 6.8 & 15.8 & 57.0 & 58.3 & 46.7 & 8.6 & 55.1 & 18.0 & 4.5 & 9.8 & 35.1 & 17.4 \\
				
				DADA \cite{vu2019dada} & & 
				80.0 & 33.8 & 75.0 & \xmark & \xmark & \xmark & 8.0 & 9.4 & 62.1 & 76.3 & 49.7 & 14.3 & 76.3 & 27.8 & 5.2 & 31.7 & 42.3 & 7.5 \\
				
				MaxSquare \cite{maxsquareloss} & & 
				70.1 & 23.3 & 72.8 & \xmark & \xmark & \xmark & 6.7 & 7.2 & 60.2 & 77.6 & 48.7 & 13.8 & 63.7 & 17.4 & 3.1 & 20.1 & 37.3 & 10.9\\
				
				AdaptSegNet \cite{tsai2018learning} & & 
				79.5 & 34.7 & 76.6 & \xmark & \xmark & \xmark & 4.1 & 5.4 & 61.0 & 80.8 & 49.3 & 18.3 & 72.1 & 26.1 & 7.5 & 29.8 & 41.9 & 4.8\\
				
				PCEDA \cite{yang2020phase} & & 
				64.5 & 33.4 & \bf77.1 & \xmark & \xmark & \xmark & \bf17.6 & 16.5 & 50.1 & 81.3 & 48.9 & 24.8 & 71.9 & 25.7 & 13.3 & 41.0 & 43.6 &10.0 \\
				
				BDL \cite{li2019bidirectional} & & 
				79.2& 33.7 & 75.3 & \xmark & \xmark & \xmark & 5.6 & 8.7 & 61.1 & 80.6 & 45.0 & 21.7 & 65.7 & 26.7 & 8.5 & 24.5 & 41.2 & 10.2 \\
				
				\rowcolor{LightCyan}
				Ours & &  
				\bf88.3 & \bf40.1 & 76.9 & \xmark & \xmark & \xmark & 11.3 & 15.9 & 68.8 & \bf81.4 & 53.3 & \bf27.3 & \bf78.2 & \bf33.1 & 16.9 & 39.4 & \bf48.5 & \bf3.6\\
				\midrule
				
				FDA & \multirow{8}{*}{$0.25$} & 
				8.6 & 9.0 & 40.8 & \xmark & \xmark & \xmark & 3.9 & 7.1 & 21.5 & 51.3 & 14.5 & 6.9 & 35.3 & 5.4 & 0.0 & 14.4 & 16.8 & 35.7 \\
				
				FADA & & 
				\bf80.8 & \bf23.5 & 59.3 & \xmark & \xmark & \xmark & 1.7 & 3.7 & \bf50.6 & 15.6 & 26.2 & 0.8 & 21.2 & 6.2 & 0.3 & 2.1 & 22.5 & 30.0 \\
				
				DADA & & 
				58.0 & 11.5 & 42.7 & \xmark & \xmark & \xmark & 4.5 & 4.2 & 31.9 & 41.2 & 23.4 & 6.0 & \bf53.9 & 8.3 & 0.4 & 14.0 & 23.1 & 26.7 \\
				
				MaxSquare & & 
				70.3 & 4.6 & 53.1 & \xmark & \xmark & \xmark & 8.1 & 6.0 & 37.2 & 61.0 & 11.2 & 3.9 & 42.3 & 6.9 & 0.4 & 3.4 & 23.7 & 24.5\\
				
				AdaptSegNet & & 
				28.4 & 7.6 & 56.8 & \xmark & \xmark & \xmark & 4.4 & 2.6 & 26.4 & 62.8 & 22.5 & 9.8 & 44.2 & 8.3 & 1.1 & 10.2 & 21.9 & 24.8\\
				
				PCEDA & & 
				15.4 & 7.2 & 64.9 & \xmark & \xmark & \xmark & \bf9.3 & 9.8 & 27.0 & 71.4 & \bf35.3 & \bf13.9 & 52.0 & 12.3 & 2.2 & \bf25.4 & 26.7 &26.9 \\
				
				BDL & & 
				46.9 & 9.1 & \bf65.5 & \xmark & \xmark & \xmark & 4.0 & 5.9 & 34.7 & 68.5 & 22.7 & 12.5 & 50.7 & 10.8 & 1.2 & 12.8 & 26.6 &21.3 \\
				
				\rowcolor{LightCyan}
				Ours & &  
				74.1 & 13.1 & 58.7 & \xmark & \xmark & \xmark & 7.7 & \bf15.1 & 39.2 & \bf73.4 & 22.0 & 12.0 & 47.2 & \bf14.0 & \bf2.3 & 21.5 & \bf30.8 & \bf20.7\\
				\midrule
				
				FDA & \multirow{8}{*}{$0.5$} &
				0.0 & 0.0 & 7.2 & \xmark & \xmark & \xmark & 1.3 & 0.7 & 17.8 & 13.7 & 0.0 & 0.0 & 2.5 & 0.2 & 0.0 & 0.0 & 3.3 & 49.2 \\
				
				FADA & & 
				\bf76.0 & \bf15.9 & \bf56.3 & \xmark & \xmark & \xmark & 0.2 & 0.6 & \bf45.0 & 0.2 & 7.6 & 0.0 & 5.2 & 0.9 & 0.0 & 0.1 & \bf16.0 & \bf36.5 \\
				
				DADA & & 
				42.9 & 2.3 & 16.3 & \xmark & \xmark & \xmark & 1.8 & 0.7 & 24.1 & 12.5 & 2.5 & 0.8 & \bf23.5 & 2.1 & 0.0 & 4.8 & 10.3 & 39.5 \\
				
				MaxSquare & & 
				42.7 & 0.2 & 25.3 & \xmark & \xmark & \xmark & \bf5.0 & 2.7 & 24.5 & 18.0 & 0.8 & 0.1 & 15.0 & 1.5 & 0.0 & 0.2 & 10.5 & 37.7\\	
				
				AdaptSegNet & & 
				2.1 & 0.4 & 24.5 & \xmark & \xmark & \xmark & 2.1 & 0.5 & 19.2 & 21.4 & 1.4 & 2.2 & 11.7 & 1.7 & 0.1 & 2.5 & 6.9 & 39.8\\
				
				PCEDA & & 
				0.1 & 0.1 & 40.0 & \xmark & \xmark & \xmark & 2.4 & 1.8 & 21.0 & 37.2 & \bf13.1 & 1.3 & 9.3 & \bf2.5 & \bf0.7 & 1.6 & 10.1&43.5 \\
				
				BDL & & 
				2.8 & 0.7 & 32.1 & \xmark & \xmark & \xmark & 2.0 & 1.8 & 20.3 & \bf53.7 & 2.7 & 1.3 & 22.3 & 1.4 & 0.4 & 1.7 & 11.0 &40.4 \\
				
				\rowcolor{LightCyan}
				Ours & &  
				15.3 & 2.0 & 26.0 & \xmark & \xmark & \xmark & 2.5 & \bf5.9 & 20.7 & 51.5 & 1.6 & \bf4.0 & 10.8 & 1.9 & 0.4 & \bf9.6 & 11.7 & 40.4\\
				
				\bottomrule
			\end{tabular}
		\end{center}
		\vspace{-0.3in}
	\end{table*}
	
	\section{Experiments}
	
	\subsection{Datasets}
	Following the same setting as previous studies, we use GTA5 \cite{richter2016playing} and SYNTHIA-RAND-CITYSCAPES \cite{ros2016synthia} as the source domain, and use Cityscapes \cite{cordts2016cityscapes} as the target domain in our study. 
	GTA5 is composed of 24,966 images (resolution: 1914 $ \times $ 1052) with pixel-accurate semantic labels, which is collected from a photo-realistic open-world game known as Grand Theft Auto V. 
	SYNTHIA-RAND-CITYSCAPES dataset is generated from a virtual city, including 9,400 images (resolution: 1280 $ \times $ 760) with precise pixel-level semantic annotations. 
	Cityscapes contains 5,000 images (resolution: 2048 $ \times $ 1024), which is a large-scale street scene datasets collected from 50 cities. 
	There are 2,975 images for training, 500 images for validation, and 1,525 images for testing. The training set from Cityscapes is used as target domain images, and we use the validation set for performance evaluation.
	
	\subsection{Implementation Details}
	\paragraph{Network Architecture}
	We follow the same experimental protocol in this area, which uses two network architectures: DeepLab-v2 \cite{chen2018deeplab} with VGG16 \cite{simonyan2014very} backbone, and DeepLab-v2 with ResNet101 backbone.
	The domain discriminator has 5 convolution layers with kernel 4$ \times $4 and stride of 2, each of which is followed by a leaky ReLU parameterized by 0.2 except the last one.
	The channel number of each layer is \{64, 128, 256, 512, 1\}. 
	
	\vspace{-10pt}
	\paragraph{Model Training}
	Adam optimizer with initial learning rate 1e-4 and momentum (0.9, 0.99) is used in DeepLab-VGG16. 
	We apply step decay to the learning rate with step size 30000 and drop factor 0.1. 
	Stochastic Gradient Descent optimizer with momentum 0.9 and weight decay 5e-4 is used in DeepLab-ResNet101. 
	The learning rate of DeepLab-ResNet101 is initialized as 1e-4 and is decreased by the polynomial policy with a power of 0.9. 
	Adam optimizer with momentum (0.9, 0.99) and initial learning rate 1e-6 is used in the domain discriminator.
	We set $\epsilon_m=1.0$ in equation \ref{eq:4} and \ref{eq:5}.
	
	\vspace{-10pt}
	\paragraph{Perturbed Test Data}
	To evaluate the robustness of our approach and existing UDA methods against adversarial attacks, we generate the perturbed image for each clean test image from Cityscapes.
	we use PSPNet \cite{zhao2017pyramid} as the surrogate model owing to its popularity in the semantic segmentation area.
	To attack PSPNet, we use FGSM with three different $\epsilon$ values, \ie, $\epsilon=0.1$, $\epsilon=0.25$, and $\epsilon=0.5$.
	For each $\epsilon$, we generate its corresponding adversarial examples and use them to evaluate the robustness of recently published UDA methods.
	For a fair comparison, we directly download the pre-trained models from the original paper and evaluate these models on the perturbed test data generated by PSPNet.
	
	\begin{figure*}
		\begin{center}
			\includegraphics[width=1.0\linewidth]{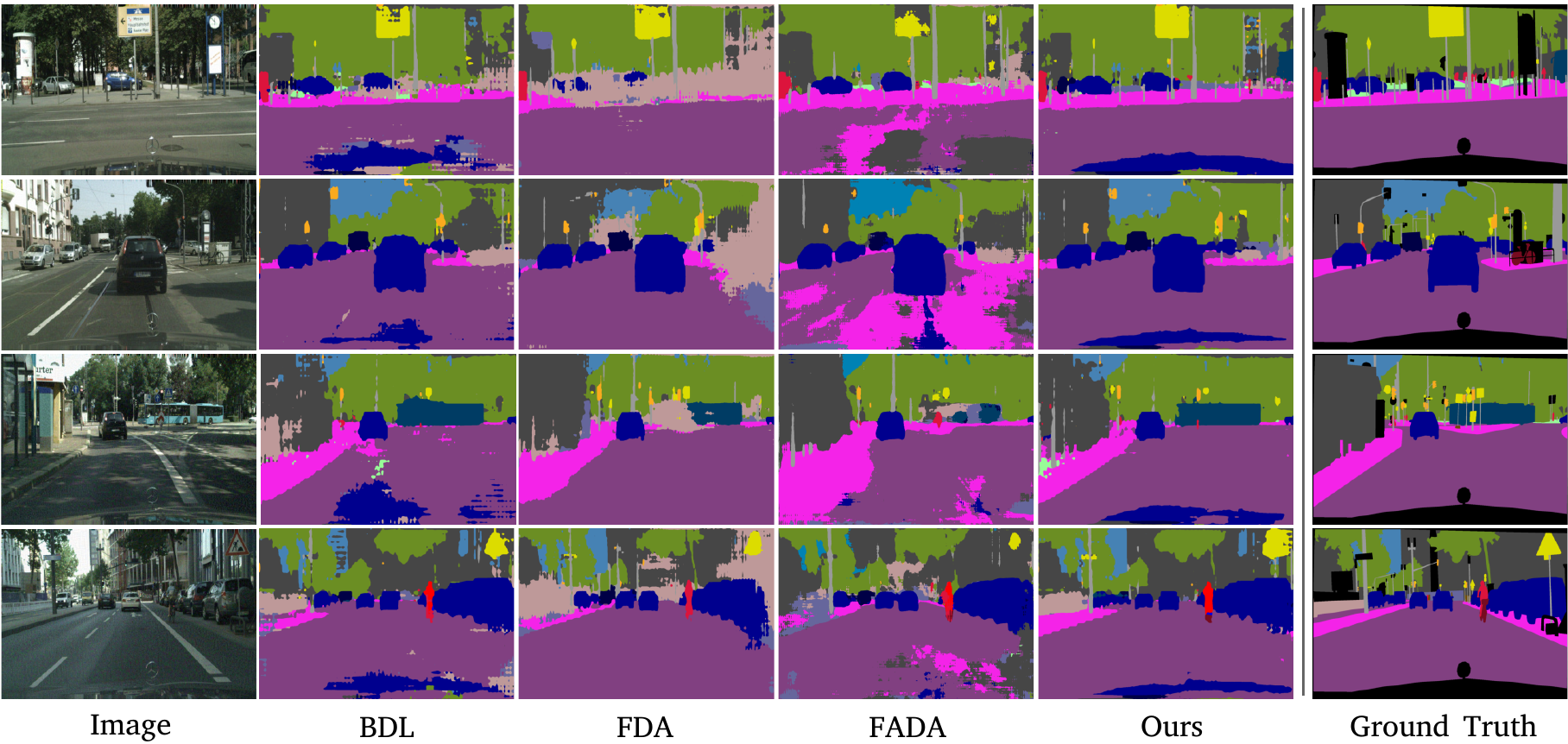}
		\end{center}
		\caption{Qualitative comparison of our method against BDL \cite{li2019bidirectional}, FDA \cite{yang2020fda}, and FADA \cite{wang2020classes} under the perturbation with $\epsilon=0.1$. The first column indicates the perturbed test images.}
		\label{fig:quali}
		\vspace{-0.2in}
	\end{figure*}
	
	\subsection{Experimental Results}
	Since the robustness of existing UDA methods remains unexplored, we first comprehensively evaluate their robustness against adversarial attacks in this section (Table~\ref{table:gta2city} and Table~\ref{table:synthia2city}). 
	We then perform a comparison of our method on two widely used benchmark settings, \ie, "GTA5 to Cityscapes" and "SYNTHIA to Cityscapes".  
	Two criteria, \ie, mIoU and mIoU drop are used for performance assessment.
	Specifically, mIoU indicates the mean IoU on perturbed test data, while mIoU drop indicates the performance degradation compared to the mIoU on clean test data.
	Therefore, the higher the mIoU and the lower the mIoU drop, the more robust the model.
	
	\vspace{-11pt}
	\paragraph{GTA5 to Cityscapes}
	As shown in Table~\ref{table:gta2city}, we achieve the best performance on all three kinds of adversarial attacks.
	In particular, even slight adversarial perturbations can mislead AdaptSegNet \cite{tsai2018learning} and BDL \cite{li2019bidirectional} and dramatically degrade their performance.
	For instance, when evaluated with VGG16 backbone on perturbed test data from $\epsilon=0.25$, they only achieve mIoU 7.9 and mIoU 14.6, with mIoU drop 27.1 and 26.7, respectively.
	By contrast, our method still gets mIoU 32.8 and only has a performance drop of mIoU 7.3.
	Similarly, two recently proposed UDA methods, \ie, FDA \cite{yang2020fda} and PCEDA \cite{yang2020phase} suffer from mIoU drop of 31.4 and 30.8, respectively. 
	The results suggest that existing UDA methods in semantic segmentation are broadly vulnerable to adversarial attacks, even a slight but intentional perturbation leads to dramatic performance degradation.
	The reason is that although these methods demonstrate remarkable performance on clean test data, none of them, however, take the adversarial attack into account.
	Instead, we innovatively propose adversarial self-supervision to take advantage of adversarial examples, to improve the robustness of UDA models.
	Note, different from previous contrastive learning models \cite{chen2020simple}, our method is tailored to UDA environments, which can adapt domain knowledge simultaneously.
	This is evident in the qualitative comparison in Figure~\ref{fig:quali}, where our method demonstrates accurate predictions on the target domain.

	\begin{figure}
		\begin{center}
			\includegraphics[width=1.0\linewidth]{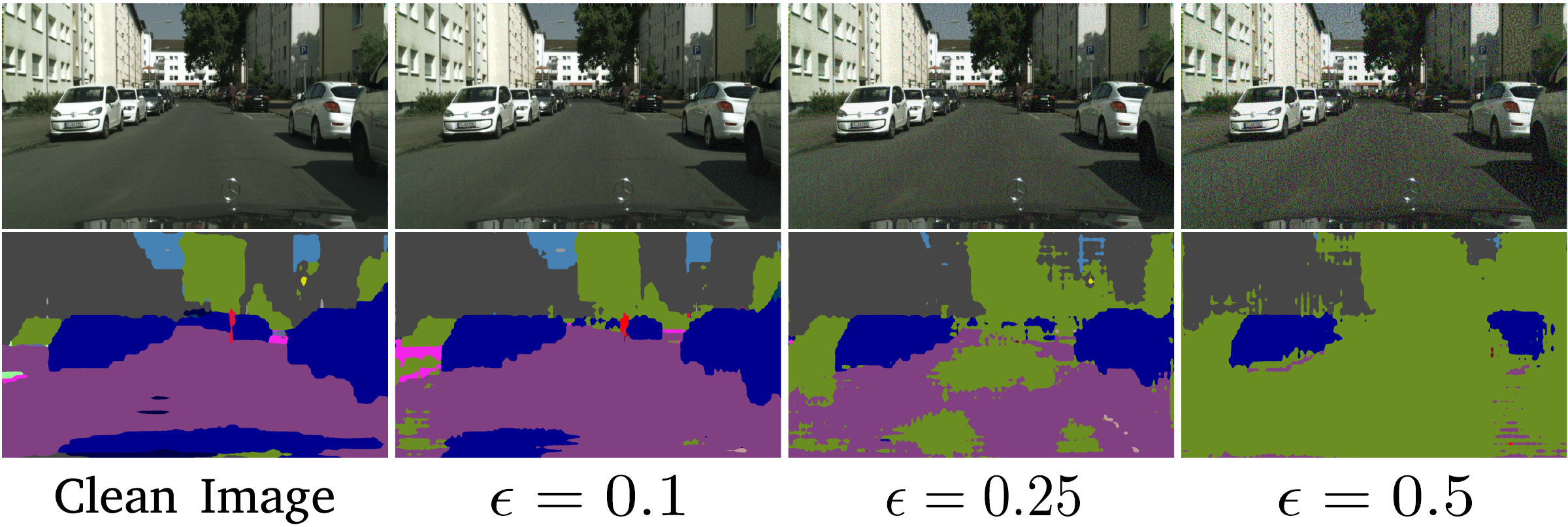}
		\end{center}
		\caption{Qualitative study of our method under three adversarial attacks, \ie, $\epsilon=0.1$, $\epsilon=0.25$, and $\epsilon=0.5$.}
		\label{fig:attack_ours}
	\end{figure}

	\vspace{-11pt}
	\paragraph{SYNTHIA to Cityscapes}
	Table~\ref{table:synthia2city} shows the performance comparison on "SYNTHIA to Cityscapes", where our method again demonstrates significant robustness improvement. 
	In contrast, other UDA methods can be easily fooled by small perturbation on the test data.
	The only exception is when evaluated with ResNet101 backbone on perturbed test data $\epsilon=0.5$, where FADA \cite{wang2020classes} achieves the best accuracy.
	Much of this interest is attributed to its fine-grained adversarial strategy for class-level alignment.
	Another interesting observation is that our model achieves better performance on perturbed test data ($\epsilon=0.1$) than on clean test data by using the VGG16 backbone.
	This can be explained by the fact that training on adversarial examples can regularize the model somewhat, as reported in \cite{goodfellow2014explaining,szegedy2013intriguing}.
	We further perform a qualitative study of our method when evaluated on the different magnitude of perturbations as shown in Figure~\ref{fig:attack_ours}. 
	
	\vspace{-10pt}
	\paragraph{Ablation Study}
	We perform the ablation study of the perturbation magnitude $\epsilon_m$ (equation \ref{eq:4} and \ref{eq:5}).
	Table~\ref{table:ablation_gta2city} reveals that $\epsilon_m=1$ results in more robust UDA model than $\epsilon_m=0.1$.
	The reason is that the adversarial examples generated by $\epsilon_m=1$ are highly perturbed compared to the adversarial examples from $\epsilon_m=0.1$, which in turn encourages our model to be more robust against perturbations.
	
	\newcommand{\cmark}{\ding{51}}%
	\begin{table}
		\caption{Ablation study of $\epsilon_m$ in our model.}
		\label{table:ablation_gta2city}
		
		\footnotesize
		\setlength\tabcolsep{8pt}
		\begin{center}
			\begin{tabularx}{.45\textwidth}{ c|c|c|c|c @{} }
				\toprule
				\multicolumn{1}{ c }{} &
				\multicolumn{2}{ c }{\bf GTA5 to Cityscapes } &
				\multicolumn{2}{ c }{\bf SYNTHIA to Cityscapes } \\
				\midrule
				$\epsilon$ & $\epsilon_m=0.1$ & $\epsilon_m=1$ & $\epsilon_m=0.1$ & $\epsilon_m=1$ \\
				\midrule
				0.1 & 35.9 & 39.4 & 37.2 & 40.9 \\
				0.25 & 21.2 & 32.8 & 21.5 & 33.6 \\
				0.5 & 8.3 & 19.1 & 8.9 & 18.5 \\
				\midrule
				\midrule
				0.1 & 42.7  & 44.6  & 45.5 & 48.5 \\
				0.25 & 30.2  & 33.1 & 26.2 & 30.8 \\
				0.5 & 14.9  & 17.6 & 10.4 & 11.7 \\
				\bottomrule
			\end{tabularx}
		\end{center}
		\vspace{-0.2in}
	\end{table}
	
	\vspace{-10pt}
	\section{Conclusion}
	In this paper, we introduce a new unsupervised domain adaptation framework for semantic segmentation.
	This is motivated by the observation that the robustness of semantic adaptation methods against adversarial attacks has not been investigated.
	For the first time, we perform a comprehensive evaluation of their robustness and propose the adversarial self-supervision by maximizing agreement between clean samples and their adversarial examples.
	Extensive empirical studies are performed to explore the benefits of our method in improving the model's robustness on adversarial attacks.
	The superiority of our method is also thoroughly proved on "GTA5-to-Cityscapes" and "SYNTHIA-to-Cityscapes".
	
	{\small
		\bibliographystyle{ieee_fullname}
		\bibliography{egbib}

\begin{thebibliography}{10}\itemsep=-1pt

\bibitem{hoffman2017cycada}
Cycada: Cycle consistent adversarial domain adaptation.
\newblock In {\em International Conference on Machine Learning (ICML)}, 2018.

\bibitem{biggio2012poisoning}
Battista Biggio, Blaine Nelson, and Pavel Laskov.
\newblock Poisoning attacks against support vector machines.
\newblock {\em International conference on machine learning (ICML)}, 2012.

\bibitem{chen2018deeplab}
Liang-Chieh Chen, George Papandreou, Iasonas Kokkinos, Kevin Murphy, and Alan~L
  Yuille.
\newblock Deeplab: Semantic image segmentation with deep convolutional nets,
  atrous convolution, and fully connected crfs.
\newblock {\em IEEE transactions on pattern analysis and machine intelligence
  (TPAMI)}, 40(4):834--848, 2018.

\bibitem{chen2020simple}
Ting Chen, Simon Kornblith, Mohammad Norouzi, and Geoffrey Hinton.
\newblock A simple framework for contrastive learning of visual
  representations.
\newblock {\em International Conference on Machine Learning (ICML)}, 2020.

\bibitem{chen2018road}
Yuhua Chen, Wen Li, and Luc Van~Gool.
\newblock Road: Reality oriented adaptation for semantic segmentation of urban
  scenes.
\newblock In {\em Proceedings of the IEEE Conference on Computer Vision and
  Pattern Recognition (CVPR)}, pages 7892--7901, 2018.

\bibitem{cordts2016cityscapes}
Marius Cordts, Mohamed Omran, Sebastian Ramos, Timo Rehfeld, Markus Enzweiler,
  Rodrigo Benenson, Uwe Franke, Stefan Roth, and Bernt Schiele.
\newblock The cityscapes dataset for semantic urban scene understanding.
\newblock In {\em Proceedings of the IEEE Conference on Computer Vision and
  Pattern Recognition (CVPR)}, pages 3213--3223, 2016.

\bibitem{doersch2015unsupervised}
Carl Doersch, Abhinav Gupta, and Alexei~A Efros.
\newblock Unsupervised visual representation learning by context prediction.
\newblock In {\em Proceedings of the IEEE international conference on computer
  vision (ICCV)}, 2015.

\bibitem{dosovitskiy2015discriminative}
Alexey Dosovitskiy, Philipp Fischer, Jost~Tobias Springenberg, Martin
  Riedmiller, and Thomas Brox.
\newblock Discriminative unsupervised feature learning with exemplar
  convolutional neural networks.
\newblock {\em IEEE transactions on pattern analysis and machine intelligence
  (PAMI)}, 2015.

\bibitem{eykholt2018robust}
Kevin Eykholt, Ivan Evtimov, Earlence Fernandes, Bo Li, Amir Rahmati, Chaowei
  Xiao, Atul Prakash, Tadayoshi Kohno, and Dawn Song.
\newblock Robust physical-world attacks on deep learning visual classification.
\newblock In {\em Proceedings of the IEEE Conference on Computer Vision and
  Pattern Recognition (CVPR)}, 2018.

\bibitem{gidaris2018unsupervised}
Spyros Gidaris, Praveer Singh, and Nikos Komodakis.
\newblock Unsupervised representation learning by predicting image rotations.
\newblock {\em International Conference on Learning Representations (ICLR)},
  2018.

\bibitem{goodfellow2014explaining}
Ian~J Goodfellow, Jonathon Shlens, and Christian Szegedy.
\newblock Explaining and harnessing adversarial examples.
\newblock {\em International Conference on Learning Representations (ICLR)},
  2015.

\bibitem{he2020momentum}
Kaiming He, Haoqi Fan, Yuxin Wu, Saining Xie, and Ross Girshick.
\newblock Momentum contrast for unsupervised visual representation learning.
\newblock In {\em Proceedings of the IEEE/CVF Conference on Computer Vision and
  Pattern Recognition (CVPR)}, 2020.

\bibitem{hendrycks2019using}
Dan Hendrycks, Kimin Lee, and Mantas Mazeika.
\newblock Using pre-training can improve model robustness and uncertainty.
\newblock {\em International Conference on Machine Learning (ICML)}, 2019.

\bibitem{hoffman2016fcns}
Judy Hoffman, Dequan Wang, Fisher Yu, and Trevor Darrell.
\newblock Fcns in the wild: Pixel-level adversarial and constraint-based
  adaptation.
\newblock {\em arXiv preprint arXiv:1612.02649}, 2016.

\bibitem{huang2020contextual}
Jiaxing Huang, Shijian Lu, Dayan Guan, and Xiaobing Zhang.
\newblock Contextual-relation consistent domain adaptation for semantic
  segmentation.
\newblock {\em Proceedings of the European Conference on Computer Vision
  (ECCV)}, 2020.

\bibitem{kim2020learning}
Myeongjin Kim and Hyeran Byun.
\newblock Learning texture invariant representation for domain adaptation of
  semantic segmentation.
\newblock In {\em Proceedings of the IEEE/CVF Conference on Computer Vision and
  Pattern Recognition (CVPR)}, pages 12975--12984, 2020.

\bibitem{li2019bidirectional}
Yunsheng Li, Lu Yuan, and Nuno Vasconcelos.
\newblock Bidirectional learning for domain adaptation of semantic
  segmentation.
\newblock In {\em Proceedings of the IEEE Conference on Computer Vision and
  Pattern Recognition (CVPR)}, 2019.

\bibitem{luc2017predicting}
Pauline Luc, Natalia Neverova, Camille Couprie, Jakob Verbeek, and Yann LeCun.
\newblock Predicting deeper into the future of semantic segmentation.
\newblock In {\em Proceedings of the IEEE International Conference on Computer
  Vision (ICCV)}, 2017.

\bibitem{luo2019taking}
Yawei Luo, Liang Zheng, Tao Guan, Junqing Yu, and Yi Yang.
\newblock Taking a closer look at domain shift: Category-level adversaries for
  semantics consistent domain adaptation.
\newblock In {\em Proceedings of the IEEE Conference on Computer Vision and
  Pattern Recognition (CVPR)}, pages 2507--2516, 2019.

\bibitem{madry2017towards}
Aleksander Madry, Aleksandar Makelov, Ludwig Schmidt, Dimitris Tsipras, and
  Adrian Vladu.
\newblock Towards deep learning models resistant to adversarial attacks.
\newblock {\em International Conference on Learning Representations (ICLR)},
  2018.

\bibitem{mei2015using}
Shike Mei and Xiaojin Zhu.
\newblock Using machine teaching to identify optimal training-set attacks on
  machine learners.
\newblock In {\em AAAI Conference on Artificial Intelligence (AAAI)}, 2015.

\bibitem{maxsquareloss}
Hongyang~Xue Minghao~Chen and Deng Cai.
\newblock Domain adaptation for semantic segmentation with maximum squares
  loss.
\newblock In {\em Proceedings of the IEEE international conference on computer
  vision (ICCV)}, 2019.

\bibitem{moosavi2016deepfool}
Seyed-Mohsen Moosavi-Dezfooli, Alhussein Fawzi, and Pascal Frossard.
\newblock Deepfool: a simple and accurate method to fool deep neural networks.
\newblock In {\em Proceedings of the IEEE conference on computer vision and
  pattern recognition (CVPR)}, 2016.

\bibitem{murez2017image}
Zak Murez, Soheil Kolouri, David Kriegman, Ravi Ramamoorthi, and Kyungnam Kim.
\newblock Image to image translation for domain adaptation.
\newblock In {\em Proceedings of the IEEE Conference on Computer Vision and
  Pattern Recognition (CVPR)}, volume~13, 2018.

\bibitem{noroozi2016unsupervised}
Mehdi Noroozi and Paolo Favaro.
\newblock Unsupervised learning of visual representations by solving jigsaw
  puzzles.
\newblock In {\em European Conference on Computer Vision (ECCV)}, 2016.

\bibitem{pan2020unsupervised}
Fei Pan, Inkyu Shin, Francois Rameau, Seokju Lee, and In~So Kweon.
\newblock Unsupervised intra-domain adaptation for semantic segmentation
  through self-supervision.
\newblock In {\em Proceedings of the IEEE/CVF Conference on Computer Vision and
  Pattern Recognition (CVPR)}, pages 3764--3773, 2020.

\bibitem{patrini2017making}
Giorgio Patrini, Alessandro Rozza, Aditya Krishna~Menon, Richard Nock, and
  Lizhen Qu.
\newblock Making deep neural networks robust to label noise: A loss correction
  approach.
\newblock In {\em Proceedings of the IEEE Conference on Computer Vision and
  Pattern Recognition (CVPR)}, 2017.

\bibitem{raju2020co}
Ashwin Raju, Chi-Tung Cheng, Yunakai Huo, Jinzheng Cai, Junzhou Huang, Jing
  Xiao, Le Lu, ChienHuang Liao, and Adam~P Harrison.
\newblock Co-heterogeneous and adaptive segmentation from multi-source and
  multi-phase ct imaging data: A study on pathological liver and lesion
  segmentation.
\newblock {\em Proceedings of the European Conference on Computer Vision
  (ECCV)}, 2020.

\bibitem{richter2016playing}
Stephan~R Richter, Vibhav Vineet, Stefan Roth, and Vladlen Koltun.
\newblock Playing for data: Ground truth from computer games.
\newblock In {\em Proceedings of the European Conference on Computer Vision
  (ECCV)}, pages 102--118, 2016.

\bibitem{ros2016synthia}
German Ros, Laura Sellart, Joanna Materzynska, David Vazquez, and Antonio~M
  Lopez.
\newblock The synthia dataset: A large collection of synthetic images for
  semantic segmentation of urban scenes.
\newblock In {\em Proceedings of the IEEE Conference on Computer Vision and
  Pattern Recognition (CVPR)}, pages 3234--3243, 2016.

\bibitem{pan2020two}
Inkyu Shin, Sanghyun Woo, Fei Pan, and In~So Kweon.
\newblock Two-phase pseudo label densification for self-training based domain
  adaptation.
\newblock {\em Proceedings of the European Conference on Computer Vision
  (ECCV)}, 2020.

\bibitem{simonyan2014very}
Karen Simonyan and Andrew Zisserman.
\newblock Very deep convolutional networks for large-scale image recognition.
\newblock In {\em International Conference on Learning Representations (ICLR)},
  2015.

\bibitem{sitawarin2018darts}
Chawin Sitawarin, Arjun~Nitin Bhagoji, Arsalan Mosenia, Mung Chiang, and
  Prateek Mittal.
\newblock Darts: Deceiving autonomous cars with toxic signs.
\newblock {\em arXiv preprint arXiv:1802.06430}, 2018.

\bibitem{su2019one}
Jiawei Su, Danilo~Vasconcellos Vargas, and Kouichi Sakurai.
\newblock One pixel attack for fooling deep neural networks.
\newblock {\em IEEE Transactions on Evolutionary Computation}, 2019.

\bibitem{sun2019unsupervised}
Yu Sun, Eric Tzeng, Trevor Darrell, and Alexei~A Efros.
\newblock Unsupervised domain adaptation through self-supervision.
\newblock {\em arXiv preprint arXiv:1909.11825}, 2019.

\bibitem{szegedy2013intriguing}
Christian Szegedy, Wojciech Zaremba, Ilya Sutskever, Joan Bruna, Dumitru Erhan,
  Ian Goodfellow, and Rob Fergus.
\newblock Intriguing properties of neural networks.
\newblock {\em International Conference on Learning Representations (ICLR)},
  2014.

\bibitem{tramer2017space}
Florian Tram{\`e}r, Nicolas Papernot, Ian Goodfellow, Dan Boneh, and Patrick
  McDaniel.
\newblock The space of transferable adversarial examples.
\newblock {\em arXiv preprint arXiv:1704.03453}, 2017.

\bibitem{tsai2018learning}
Yi-Hsuan Tsai, Wei-Chih Hung, Samuel Schulter, Kihyuk Sohn, Ming-Hsuan Yang,
  and Manmohan Chandraker.
\newblock Learning to adapt structured output space for semantic segmentation.
\newblock In {\em Proceedings of the IEEE Conference on Computer Vision and
  Pattern Recognition (CVPR)}, 2018.

\bibitem{vu2019dada}
Tuan-Hung Vu, Himalaya Jain, Maxime Bucher, Matthieu Cord, and Patrick
  P{\'e}rez.
\newblock Dada: Depth-aware domain adaptation in semantic segmentation.
\newblock In {\em Proceedings of the IEEE international conference on computer
  vision (ICCV)}, 2019.

\bibitem{wang2020classes}
Haoran Wang, Tong Shen, Wei Zhang, Lingyu Duan, and Tao Mei.
\newblock Classes matter: A fine-grained adversarial approach to cross-domain
  semantic segmentation.
\newblock {\em Proceedings of the European Conference on Computer Vision
  (ECCV)}, 2020.

\bibitem{wang2020differential}
Zhonghao Wang, Mo Yu, Yunchao Wei, Rogerio Feris, Jinjun Xiong, Wen-mei Hwu,
  Thomas~S Huang, and Honghui Shi.
\newblock Differential treatment for stuff and things: A simple unsupervised
  domain adaptation method for semantic segmentation.
\newblock In {\em Proceedings of the IEEE/CVF Conference on Computer Vision and
  Pattern Recognition (CVPR)}, pages 12635--12644, 2020.

\bibitem{xu2019self}
Jiaolong Xu, Liang Xiao, and Antonio~M L{\'o}pez.
\newblock Self-supervised domain adaptation for computer vision tasks.
\newblock {\em IEEE Access}, 2019.

\bibitem{yang2020label}
Jinyu Yang, Weizhi An, Sheng Wang, Xinliang Zhu, Chaochao Yan, and Junzhou
  Huang.
\newblock Label-driven reconstruction for domain adaptation in semantic
  segmentation.
\newblock {\em Proceedings of the European Conference on Computer Vision
  (ECCV)}, 2020.

\bibitem{yang2020phase}
Yanchao Yang, Dong Lao, Ganesh Sundaramoorthi, and Stefano Soatto.
\newblock Phase consistent ecological domain adaptation.
\newblock In {\em Proceedings of the IEEE/CVF Conference on Computer Vision and
  Pattern Recognition (CVPR)}, 2020.

\bibitem{yang2020fda}
Yanchao Yang and Stefano Soatto.
\newblock Fda: Fourier domain adaptation for semantic segmentation.
\newblock In {\em Proceedings of the IEEE/CVF Conference on Computer Vision and
  Pattern Recognition (CVPR)}, pages 4085--4095, 2020.

\bibitem{zhan2017mix}
Xiaohang Zhan, Ziwei Liu, Ping Luo, Xiaoou Tang, and Chen~Change Loy.
\newblock Mix-and-match tuning for self-supervised semantic segmentation.
\newblock {\em AAAI Conference on Artificial Intelligence (AAAI)}, 2018.

\bibitem{zhang2018fully}
Yiheng Zhang, Zhaofan Qiu, Ting Yao, Dong Liu, and Tao Mei.
\newblock Fully convolutional adaptation networks for semantic segmentation.
\newblock In {\em Proceedings of the IEEE Conference on Computer Vision and
  Pattern Recognition (CVPR)}, pages 6810--6818, 2018.

\bibitem{zhang2018generalized}
Zhilu Zhang and Mert Sabuncu.
\newblock Generalized cross entropy loss for training deep neural networks with
  noisy labels.
\newblock In {\em Advances in neural information processing systems (NeurIPS)},
  2018.

\bibitem{zhao2017pyramid}
Hengshuang Zhao, Jianping Shi, Xiaojuan Qi, Xiaogang Wang, and Jiaya Jia.
\newblock Pyramid scene parsing network.
\newblock In {\em Proceedings of the IEEE Conference on Computer Vision and
  Pattern Recognition (CVPR)}, pages 2881--2890, 2017.

\bibitem{zhu2018penalizing}
Xinge Zhu, Hui Zhou, Ceyuan Yang, Jianping Shi, and Dahua Lin.
\newblock Penalizing top performers: Conservative loss for semantic
  segmentation adaptation.
\newblock In {\em Proceedings of the European Conference on Computer Vision
  (ECCV)}, pages 568--583, 2018.

\bibitem{zou2018unsupervised}
Yang Zou, Zhiding Yu, BVK~Vijaya Kumar, and Jinsong Wang.
\newblock Unsupervised domain adaptation for semantic segmentation via
  class-balanced self-training.
\newblock In {\em Proceedings of the European Conference on Computer Vision
  (ECCV)}, pages 297--313. Springer, 2018.

\end{thebibliography}
	}
	
\end{document}